\documentclass[runningheads]{llncs}
\usepackage{url}
\usepackage{amsmath}
\usepackage[mathscr]{euscript}
\usepackage{amssymb}
\usepackage{mathtools}
\usepackage{stmaryrd}
\usepackage{bm}
\usepackage{courier}
\usepackage{cite}
\usepackage[breaklinks=true]{hyperref}
\usepackage{breakcites}
\DeclareMathOperator*{\argmin}{arg\,min}
\DeclareMathOperator*{\argmax}{arg\,max}

\title{Neighborhood Sensitive Mapping for Zero-Shot Classification using Independently Learned Semantic Embeddings}

\titlerunning{Neighborhood Sensitive Mapping for Zero-Shot Classification}

\author{Gaurav Singh\inst{1},
Fabrizio Silvestri\inst{2},
John Shawe-Taylor\inst{1}\\
}
\authorrunning{G. Singh et al.}
\institute{University College London\\
\email{\{g.singh\}@cs.ucl.ac.uk} \\
\email{\{j.shawe-taylor\}@ucl.ac.uk} \and
Yahoo\\
\email{\{silvestr\}@yahoo-inc.com}}

\begin{document}
\maketitle
\begin{abstract}
In a traditional setting, classifiers are trained to approximate a target function $f:X \rightarrow Y$ where at least a sample for each $y \in Y$ is presented to the training algorithm. In a zero-shot setting
we have a subset of the labels $\hat{Y} \subset Y$ for which we do not observe any corresponding training instance. Still, the function $f$ that we train must be able to correctly assign labels also on $\hat{Y}$. In practice, zero-shot problems are very important especially when the label set is large and the cost of editorially label samples for all possible values in the label set might be prohibitively high. Most recent approaches to zero-shot learning are based on finding and exploiting relationships between labels using semantic embeddings.
 We show in this paper that semantic embeddings, despite being very good at capturing relationships between labels, are not very good at capturing the relationships among labels in a data-dependent manner. For this reason, we propose a novel two-step process for learning a zero-shot classifier. In the first step, we learn what we call a \emph{property embedding space} capturing the ``\emph{learnable}'' features of the label set. Then, we exploit the learned properties in order to reduce the generalization error for a linear nearest neighbor-based classifier.

\end{abstract}

\section{Introduction}
One of the most prominent areas of research in machine learning is \emph{classification}. In a traditional setting, the problem of classification consists of training a model to approximate a target function $f:\mathcal{X} \rightarrow \mathcal{Y}$ where at least a sample for each $y \in \mathcal{Y}$ is presented to the training algorithm. A common problem faced by many systems, especially in their early stage is the absence of training instances for all possible classes. In such cases, a traditional classifier cannot, in fact, assign class labels that have not been observed in the training set. This is one of the main reasons for the growth of a research area, \emph{zero-shot classification}, studying how classification can be done also using unseen labels. In a zero-shot setting  ~\cite{palatucci2009zero}, we are given a subset $\hat{\mathcal{Y}} \subset \mathcal{Y}$ of the labels for which we do not observe any corresponding training instance. Still, the function $f$ that we train must be able to correctly assign labels also on $\hat{\mathcal{Y}}$. There is a growing need for using classification systems in order to automate different online and offline tasks, often having labels that are yet to be observed in the training data.
Based on the definitions given by Palatucci \emph{et al.} \cite{palatucci2009zero}, we address the following general research question: ``\emph{Given a semantic encoding of a large set of concept labels, can we build a classifier to recognize labels that were omitted from the training set?}''

To correctly deal with unseen labels, one possibility is to establish relationship between seen and unseen labels. One of the earliest works in zero-shot classification \cite{palatucci2009zero} proposes a method based on creating semantic embeddings for words based on co-occurrences of labels in dictionaries and human feedback on label properties. More recently, the progress made in learning semantic embeddings (e.g. word2vec \cite{mikolov2013efficient}) has provided a method for encoding the semantic meaning of words. Therefore, classification tasks where the label set is made up of meaningful words can be used to establish such inter-label relationships. An input is mapped to the label embedding space, and then, a nearest neighbors approach is used to predict the correct label, including from those not seen in the training set. Unfortunately, these methods that are aimed at ``directly learning'' a mapping from the input to a semantic space may suffer from two major problems. 

First, the target semantic space may be hard to learn due to the noise. This is because similarity between word embeddings of two labels may not correspond to the similarity between their respective inputs. Such a situation can occur because the word embeddings for labels were constructed using an independent dataset (i.e. google news).  As an example, if two words (labels) occur together in google news corpora - that was used to learn the word embeddings - then they would have similar embedding representation, with no regards to their similarity in the input space. Hence, semantic space representations (i.e. word embeddings) for labels can vary significantly based on the dataset used to learn them, and this could be a problem in zero-shot classification. 


Second, nearest neighbor classifiers for zero-shot classification are not designed to take into account the neighborhood of a given label in the embedding space. This basically means that labels that have really close neighbors should be learned with more accuracy at the expense of labels that are isolated in their neighborhood.

We address the two above mentioned problems in a step-wise manner. Firstly, we need to fine-tune label embeddings based on the task. Secondly, we need to develop a neighborhood sensitive mapping that can reduce the risk of error for a nearest neighbor classifier during zero-shot classification. 

Our main contributions in this study are:
\begin{enumerate}
    \item Learning data-dependent embeddings for labels
    \item Neighborhood sensitive mapping to reduce classification error
\end{enumerate}
 
\section{Related Works}
There has been active research in the area of zero-shot classification in the recent past. Originally, the problem was defined in its current form by Palatucci \emph{et al.} \cite{palatucci2009zero}, where they address the problem by embedding the set of labels into a semantic space used then to extrapolate information about unseen labels. Given an object in the dataset, they firstly predict the set of semantic features (in the embedding space) corresponding to that input, and then they find the nearest class in the labels embedding. They also develop a generalization bound on the error for zero-shot classification using a nearest neighbour classifier. A following work \cite{hariharan2012efficient} proposed a multi-label max-margin classifier with applications to zero-shot classification. By means of a correlation matrix between different labels they are able to predict unseen labels. The method, specifically designed for multi-label classification, explicitly reduces the hamming loss between prediction vectors and the label vectors.  

There are numerous practical applications where unseen images need to be classified. As a result, zero-shot classification has found considerable interest in the area of image recognition and classification. Another such work  \cite{lampert2014attribute} develops an approach specifically based on learning attribute for animals (e.g. color, eating habits etc.). These attributes are not unique to a single animal, and therefore, can be learned from the available training data. One more work \cite{mensink2014costa} proposed a method specifically targeting images, that focuses on exploiting co-occurrences of visual concepts in images for knowledge transfer. At the same time, Jayaraman \emph{et al.} \cite{jayaraman2014zero} propose an approach for zero-shot classification, for when image attributes are unreliable, and use the error tendencies of the different attributes to develop a linear discriminant model. There have been many such attribute based learning methods for visual recognition that have been developed in the past ~\cite{lampert2009learning, farhadi2009describing, yu2013designing, rohrbach2011evaluating, turakhia2013attribute, parikh2011relative}, to point to a few.

Unfortunately, there are situations when such attribute information is not easily available, and therefore, we need to build a semantic space for labels. Building such a space may lead to an additional overhead and it may also require an extensive knowledge of the domain in which labels are defined. For these reasons, the learned semantic embeddings may not even be of the required high quality. To address these concerns, a general methodology to learn word embeddings is presented by Mikolov \emph{et al.} \cite{mikolov2013efficient}. The technique, known as \emph{word2vec}, has made a advances in both efficiency and quality of the vectors learned. These vector representations are easily available for billions of words trained on terabytes of \textit{Google News} and \textit{Wikipedia} data. By using semantic embeddings learned using word2vec, Norouzi \emph{et al.} \cite{norouzi2013zero} proposed \emph{ConSE}, a zero-shot image classification system specifically tested on image classification. ConSE uses a convolution network to embed an image into a vector space and uses a convex combination of nearest label embeddings to construct the prediction vector. It assumes that the set of predicted labels is disjoint to the set of seen labels, an assumption that is neither valid in the real world nor in our work.  Another work \cite{romera2015embarrassingly} proposed a linear regression model for zero-shot classification that also relies on independently learned semantic embeddings. This work is very similar in practice to the model of Palatucci \emph{et al.} \cite{palatucci2009zero}. Another work \cite{Li_2015_ICCV} proposed to learn semantic embeddings for labels from scratch without using independently learned semantic embeddings, which differentiates it from our work.

An interesting work based on semi-supervised learning \cite{li2015max} proposed a max-margin multi-class zero-shot classifier with the assumption that unlabeled data is available during training, an assumption that is not  made in the paper defining the problem of zero-shot classification and that, therefore, we are not making in this chapter as well. One very recent work \cite{zhang2015zero} proposes to learn semantic similarity embeddings for zero-shot classification,  such that, each source or target label is represented as a mixture of seen label proportions. The method does well to predict unseen labels when the set of possible labels does not include seen labels, since the model is biased towards predicting seen labels. But, the method does not perform as well in the real world scenarios where it is not possible to distinguish between an unseen and a seen label at test time. Contrary to this, we work on a method that focuses on learning an accurate mapping from input data to the semantic space representation of labels, which leads to better accuracy in predicting the correct label.

\section{Proposed Method}
In this section, we present our two step approach to zero-shot classification. We first describe the method to learn the proposed data-dependent embeddings that we refer to as \textit{property embeddings}. Afterwards, we present how to minimize the classification error for nearest neighbour zero-shot multi-class classification using these property embeddings.



\subsection{Property Embeddings}
\label{sec:learn_properties}
Most previous methods either assume the presence of reliable attribute information for labels \cite{romera2015embarrassingly} or directly use noisy word embeddings \cite{norouzi2013zero}. While some other methods make assumptions about the availability of unlabeled data during training \cite{li2015max}, when in fact, this is often not possible in the real world. On the other hand, semantic embeddings like word2vec \cite{mikolov2013efficient} and Glove \cite{pennington2014glove}, or more recently FastText \cite{bojanowski2017enriching} - while easy to obtain - are generally learned independently of the task at hand. Consequently, these embeddings are noisy, and therefore, can lead to inferior performance on the task at hand. Hence, we attempt to re-learn the label embeddings (obtained using word2vec on google news) \emph{explicitly} using task data, and refer to them as property embeddings.

There is a one-to-one correspondence between labels and their property embeddings i.e. each label has a label embedding and a property embedding. These property embeddings encode labels such that: 1) the new property embeddings can be learnt from the input; 2) the similarities among labels in the label embeddings space are preserved in the property embedding space as much as possible. This leads us to formulate the objective $J_s$ in Equation (\ref{eq:j11}) in which the first part ensures that the matrix of property embeddings $B$ can be mapped from data $X$ using the model $W$, while the second part ensures that the cosine similarities that existed between different labels in the original label embeddings are preserved. We take the average of the features for all instances that have the same label, therefore, we have one input $\mathbf{x}$ for each label. We denote by $\mathcal{S}/\mathcal{U}$ the set of seen/unseen labels and the subscript $s/u$ refers to parameters corresponding to seen/unseen labels. 


\begin{equation}
J_s = \alpha \| XW-B_s \|^2 + (1-\alpha) \| B_s B_s^T - L_s L_s^T \|^2 +  \lambda \|W\|^2
\label{eq:j11}
\end{equation}
\begin{equation}
\label{eq:bs}
W, B_s = \argmin_{W, B_s} J_s
\end{equation}
Here $X \in \mathbb{R}^{|\mathcal{S}| \times d}$ is the input matrix, $B_s\in \mathbb{R}^{|\mathcal{S}| \times k_1}$ is the new property embedding matrix, $L_s\in \mathbb{R}^{|\mathcal{S}| \times k_2}$ is the label embedding matrix, $W \in \mathbb{R}^{d \times k_1}$ is a linear model, and $d,k_1, k_2$ are the dimensions of the input, property embedding and label embedding respectively. Please note that $W$ is only used to learn $B_s$, once we obtain $B_s$ we throw away the learned $W$, and re-learn its neighbourhood sensitive version in the next section. Please also note that the dimensions of property embeddings can be different from label embeddings.

We model a slightly different objective to learn property embeddings for zero-shot labels (i.e. $B_u$). The only piece of information we have about zero-shot labels is their relative similarity with other labels in the original label embedding space. This is because we do not observe these labels in the train set.  Therefore, we need to preserve the similarity between these zero-shot labels and the seen labels in the new property embedding space, leading us to the objective:

\begin{equation}
J_u =  \| B_s B_u^T- L_s L_u^T\|^2
\end{equation}
\begin{equation}
\label{eq:bu}
B_u = \argmin_{B_u} J_u
\end{equation}
Please note that $B_s$ is fixed in this objective and we use the value obtained in Equation \ref{eq:bs}. 

We optimize both objectives (Eq \ref{eq:bs} and Eq \ref{eq:bu}) using gradient descent, but we recognize that Eq \ref{eq:bu} can also be solved in closed form. We also acknowledge that there might be other methods that can be applied to solve these objectives e.g. second-order methods, which might differ from gradient descent in terms of time-space performance, but that should not affect the quality of the final results.

\subsection{Neighborhood Sensitive Mapping}
\label{sec:learning_classifier}

In this section, we propose an objective for nearest neighbor classification that takes into account the neighbourhood of the label while mapping input into the property embedding space. Our aim is to learn labels that are in crowded (i.e. have a lot of labels nearer to them in the property embedding space) neighbourhood with more precision at the expense of labels that are in sparse neighbourhood. This is because a label in crowded neighbourhood is more likely to be misclassified by a nearest neighbour classifier, as compared to a label in a sparse neighbourhood, therefore, we refer to the approach as \textit{Neighborhood Sensitive Mapping}. To achieve this, we formulate the objective as:
\begin{equation}
\begin{aligned}
W =\argmax_W \sum_{(\mathbf{x}, \hat{q}) \in \mathcal{D}} \left ( \sum_{\tilde{q} \in \left(\mathcal{S} \bigcup \mathcal{U}\right) - \hat{q}}  \log\left(\sigma\left(\mathbf{x}~W \left(\mathbf{b}_{\hat{q}}-\mathbf{b}_{\tilde{q}}\right)^{T}\right)\right) \right)
-\lambda \| W \|^2
\end{aligned}
\end{equation}
Here $\mathbf{x} \in \mathbb{R}^{1 \times d}$ is the input, $\hat{q}$ is the true label, $\mathcal{S}$ is the set of seen labels, $\mathcal{U}$ is the set of unseen labels, $\mathbf{b}_q \in \mathbb{R}^{1 \times k_1}$ is the property embedding of label $q$, $\sigma$ is the sigmoid function and $W \in \mathbb{R}^{d \times k_1}$ is a linear mapping from input to property embedding space.

The first part of the above objective ensures that the dot product of the prediction vector ($\mathbf{x}W$) with the property embedding of the true label ($\mathbf{b}_{\hat{q}}$) is larger in comparison to the other labels. While the second part is the standard $l$2 regularization over the learned parameter $W$. 

\section{Experimental Evaluation}
In this section, we first describe the fMRI dataset that was used for experimentation. After that, we pose three different research questions and analyze the results in light of these questions over the fMRI dataset, exactly as in the original zero-shot paper by Palatucci \emph{et al.} \cite{palatucci2009zero}, which also analyzed these three questions over fMRI dataset, and also draw comparisons with competitive baselines. After the analysis over these research questions, we describe two additional text classification datasets. We perform final experiments on these two text tagging datasets and discuss the obtained results.


\subsection{fMRI Data}

We used the fMRI dataset for experimental analysis over three research questions, exactly as in the original paper on zero-shot \cite{palatucci2009zero}. The fMRI dataset is composed of the neural activity observed from nine human participants while looking at 60 different concrete words. These 60 words are divided into 12  categories, like animals: bear, dog, cat, cow, horse and vehicles: truck, car, train, airplane, bicycle. Each participant was shown a word and a small line drawing of the concrete object the word represents. The participants were asked to think about the properties of these objects for several seconds while scans of their brain activity were recorded. Each sample measures the neural activity at roughly 20,000 locations in the brain. Six fMRI scans were taken for each word. We also used the same time-averaging described in Mitchell \emph{et al.} \cite{mitchell2008predicting, palatucci2009zero} to create a single average brain activity pattern for each of the 60 words, for each participant. The semantic embeddings used in the experiments were 300 dimensional vectors trained on Google news using \emph{word2vec} \cite{mikolov2013efficient}, and are freely available online.

\subsection{Research Questions}
In this section we perform experimental analysis in the light of three different research questions that were posed in the original zero-shot paper \cite{palatucci2009zero}, and compared our results to LM \cite{palatucci2009zero} and ConSE \cite{norouzi2013zero}. For this analysis, we use fMRI dataset as used by Palatucci \emph{et al.} \cite{palatucci2009zero}.

We refer to our proposed method as \emph{NSM} i.e. Neighborhood Sensitive Mapping, Palatucci \emph{et al.} \cite{palatucci2009zero} is referred to as \emph{LM} and Norouzi \emph{et al.} \cite{norouzi2013zero} is referred to as \emph{ConSE}. In order to get insights into the effect of our novel property embeddings, we also test both NSM and LM with property embeddings. Therefore, \emph{NSM-PB} and \emph{LM-PB} refer to NSM using property embeddings and LM using property embeddings respectively, on the other hand, \emph{NSM} and \emph{LM} use the label embeddings.  

\textit{1. How well can the model differentiate between two novel classes, where neither class appears in the train dataset?}

We randomly draw a set of 1000 pairs of classes. We select one pair at a time and remove both of the classes in the pair from the training set, and then evaluate the accuracy of binary classification between these two removed classes on the test set. We then report the average binary classification accuracy over these pairs. 

\begin{figure}[hbt]
\centering
\includegraphics[width=13cm,height=6cm]{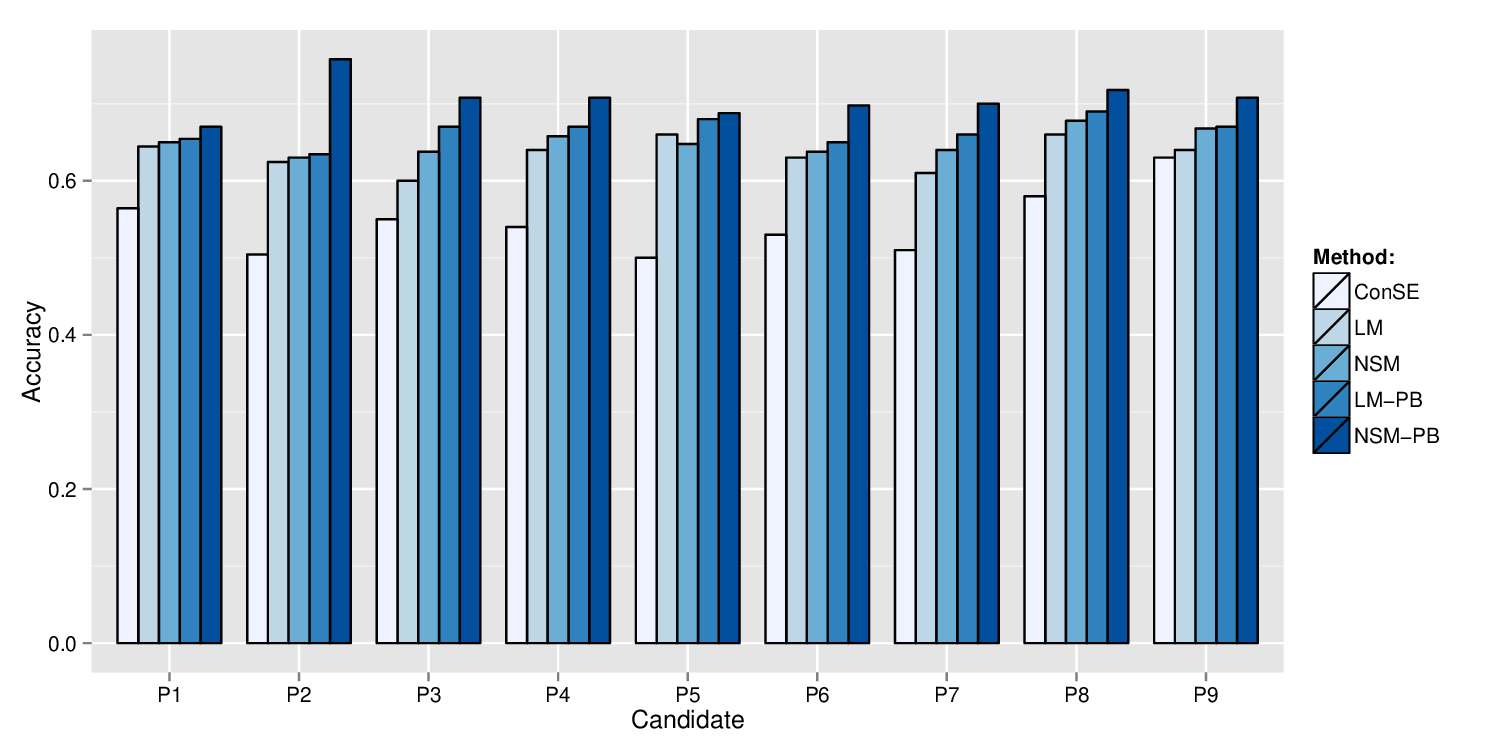}
\caption{Binary classification accuracy for different participants in fMRI dataset using our proposed approach (referred to as \textbf{NSM-PB}) versus LM-PB, LM, NSM and ConSE. The results are averaged over 1000 different pairs of binary classes and the results are statistically significant at p-value = 0.0016 using a two sided t-test. 
The average results for NSM-PB, LM-PB, ConSE, LM and NSM: 0.7049, 0.6633, 0.5454, 0.6420 and 0.6495 respectively. Please note that the classification accuracy of a random classifier would be 0.5.}
\label{fig:binary_classification}
\end{figure}

We can see that NSM-PB outperforms LM-PB, LM (Figure \ref{fig:binary_classification}) and ConSE. The difference between NSM-PB and others is statistically significant using a t-test at p-value $<$ 0.001. It can be seen that relearning of semantic embedding using the data, as well as modified neighborhood sensitive approach manages to better differentiate between novel classes. In fact, during our experimentation we observed that it works particularly well in case of classes that are very close in original semantic space, e.g. \textit{hammer} and \textit{chisel}. In those cases the use of a LM leads to results that are no better than random. We obtained an accuracy of 50\% for \textit{hammer} and \textit{chisel} using LM, whereas 67\% accuracy using NSM-PB.

\textit{2. How well can the model classify accurately in a multi-class classification setting, where all the test classes are absent in the train dataset?}

\begin{figure}[htb]
\begin{center}
\includegraphics[width=13cm,height=6cm]{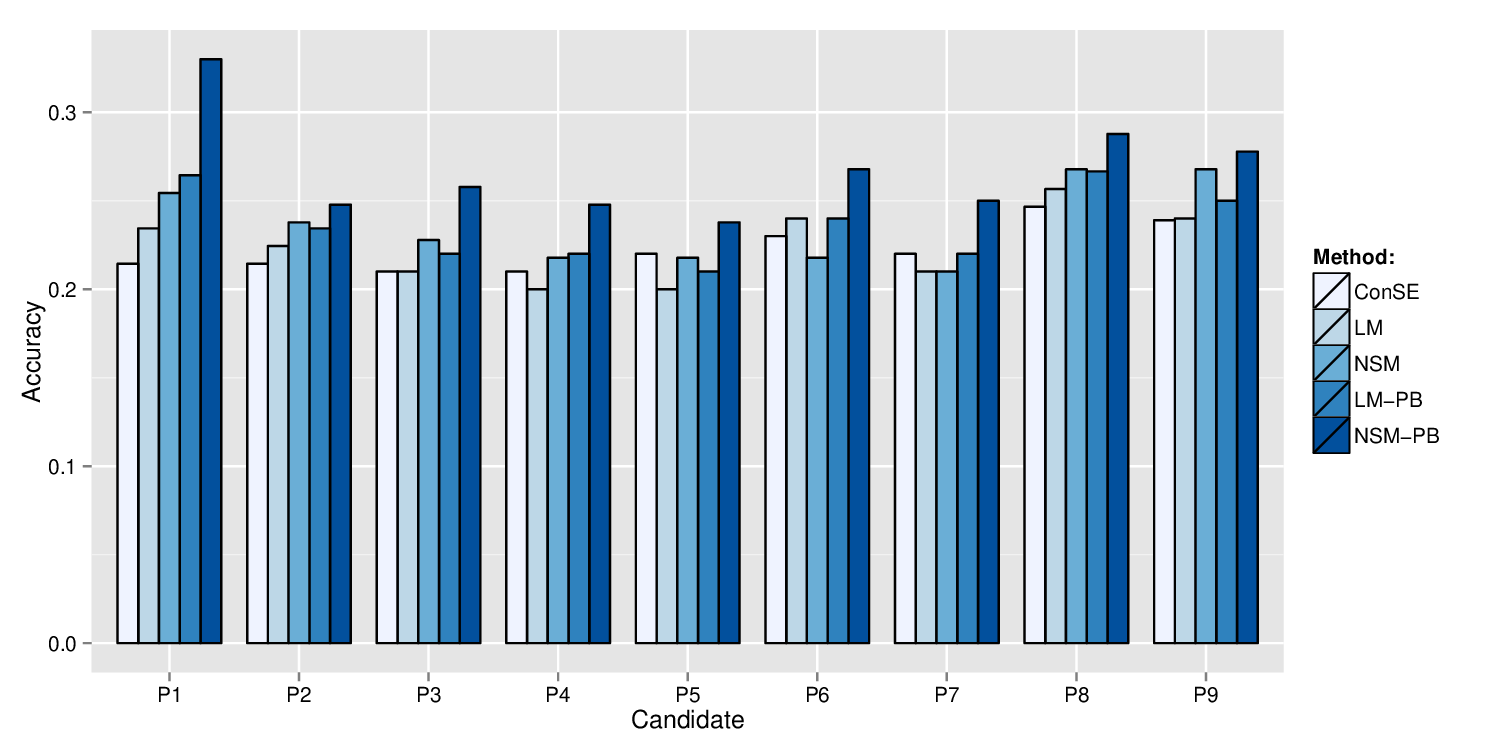}
\caption{The figure plots multi-class classification accuracy for different participants in fMRI dataset. The figure compares our proposed approach (referred to as \textbf{NSM-PB}) against LM-PB, LM, NSM and ConSE. The results are averaged over 1000 different set of five classes and the results are statistically significant at p-value = 0.0015 using a two sided t-test. In this experiment, we test the ability of NSM-PB to distinguish between five novel classes that have not been seen in the train set. The average results for NSM-PB, LM-PB, ConSE, LM and NSM are: 0.2671, 0.2356, 0.2281, 0.22393 and  0.2349 respectively.}
\label{fig:multi_classification}
\end{center}
\end{figure}

We randomly select a set of 1000 groups such that each group consists of five different classes. We compute the average of the results of multi-class classification accuracy on all these groups. We restrict the predicted class to the classes in the group, i.e. we try to predict the correct class from among the five classes in each group. The results are presented in Figure \ref{fig:multi_classification} and are statistically significant using a two sided t-test at p-value = 0.0015. Even in a multi-class setting, property embedding based models outperform the semantic embedding based models. Also, NSM-PB outperforms ConSE, even though ConSE is competitive.

\begin{figure}[htb]
\begin{center}
\includegraphics[width=13cm,height=6cm]{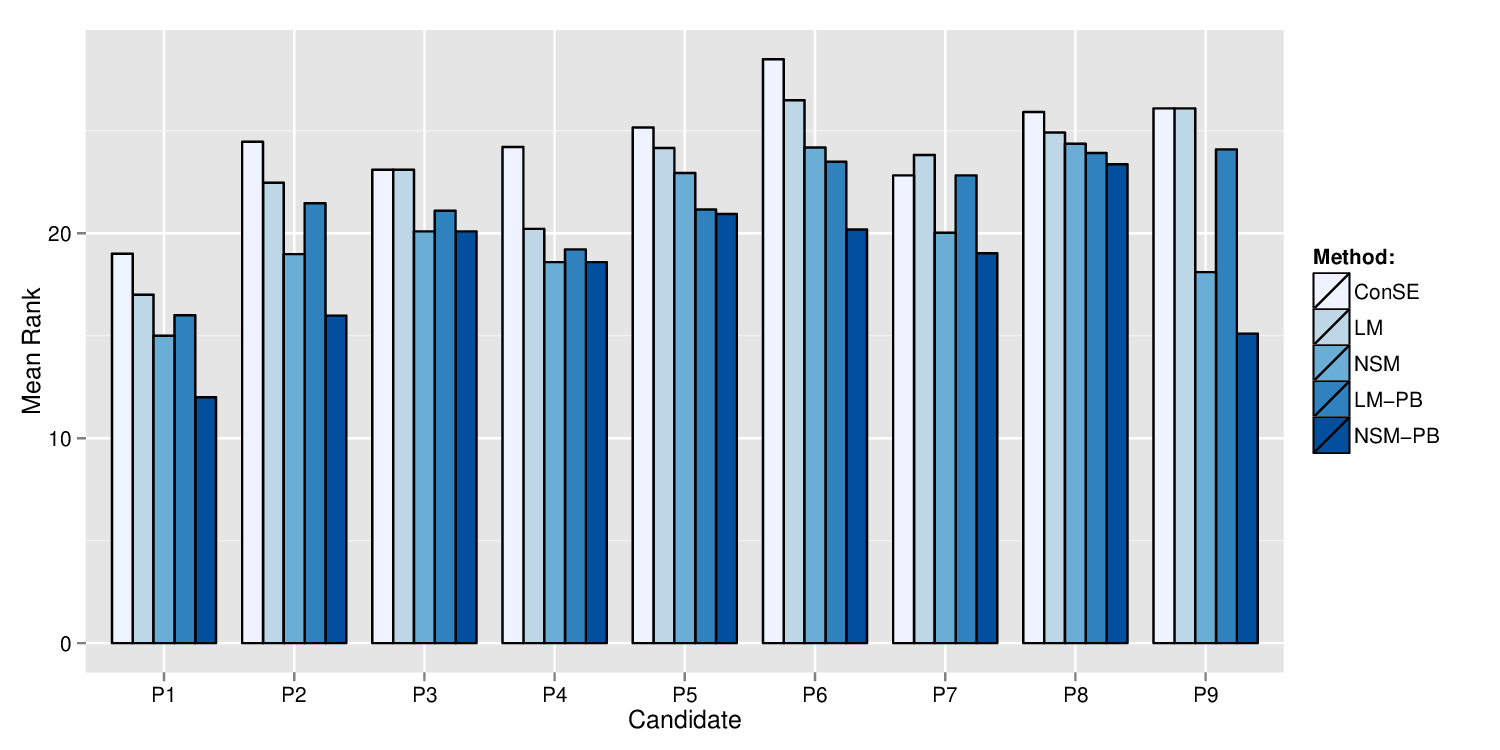}
\caption{The figure plots mean rank for a true novel class (not seen in the train set) in the ordered prediction list of classes for different participants. In the figure our proposed approach (referred to as \textbf{NSM-PB}) is compared against LM-PB, LM, NSM and ConSE. The difference of NSM-PB against LM and ConSE is statistically significant at p-value $< 0.001$ using a two sided t-test. 
The average results for NSM-PB, LM-PB, ConSE, LM and NSM are: 18.58, 21.47, 24.36, 23.137 and  20.25 respectively.}
\label{fig:mean_rank}
\end{center}
\end{figure}

We can see in Figure \ref{fig:multi_classification} that NSM-PB performs better than both LM and ConSE. It shows that the proposed model is better at discriminating between multiple zero-shot classes in a multi-class classification setting. We can see in Figure \ref{fig:mean_rank} that the mean rank of the correct label in the prediction list is much lower in case of NSM-PB as compared to both LM-PB, LM and ConSE. It shows the effectiveness of NSM-PB in predicting correct labels higher in the prediction list.

\textit{3. How well can the model predict accurately in a multi-class classification setting, where the classifier has to choose from all possible classes?}

\begin{figure}[t]
\begin{center}
\includegraphics[width=13cm,height=6cm]{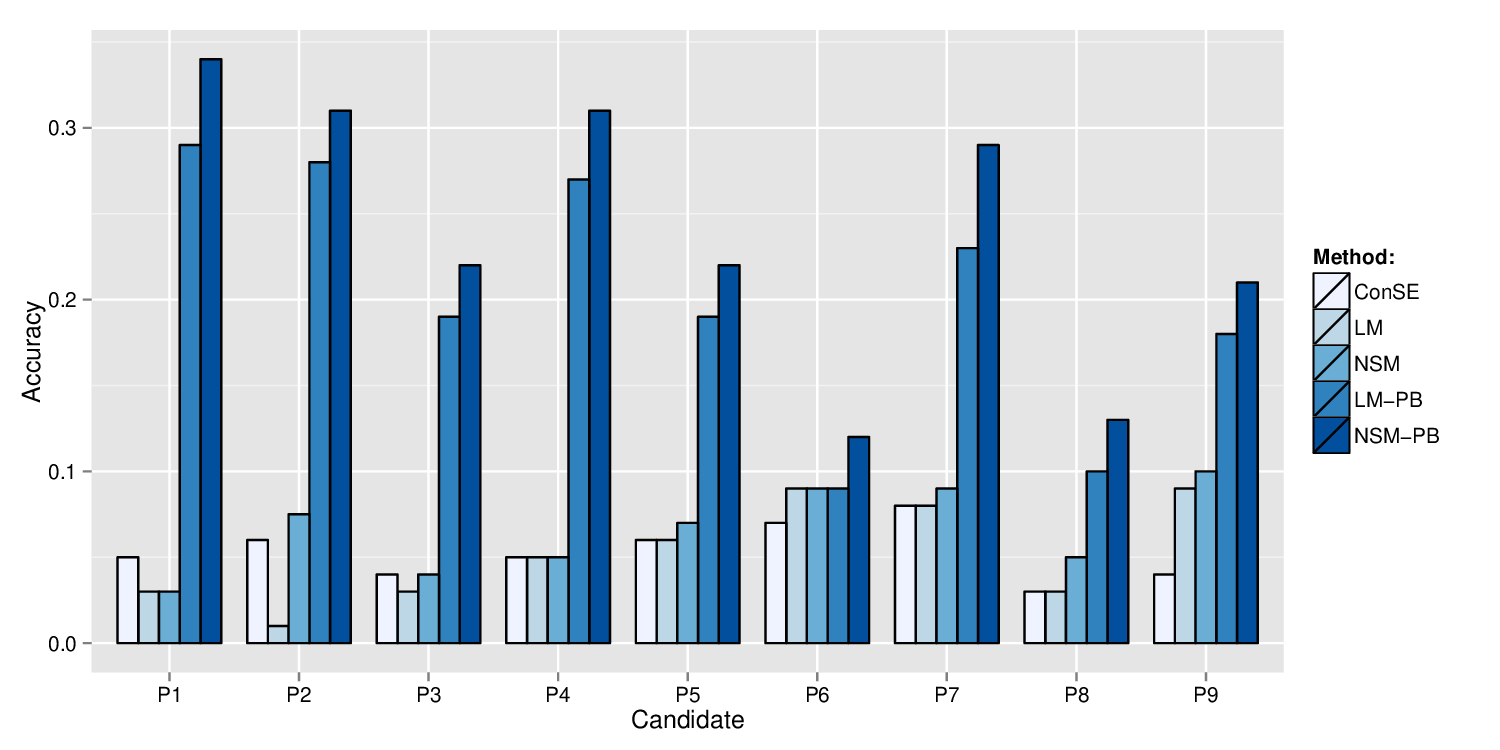}
\caption{The figure plots mean accuracy for a true novel class (not seen in the train set) for different participants in fMRI dataset. In the figure our proposed approach (referred to as \textbf{NSM-PB}) is compared against LM-PB, LM, NSM and ConSE. The difference of NSM-PB against LM and ConSE is statistically significant at p-value $< 0.001$ using a two sided t-test. 
The average results for NSM-PB, LM-PB, ConSE, LM and NSM are: 0.2389, 0.2000, 0.0533, 0.0522 and  0.0661 respectively.}
\label{fig:mean_acc}
\end{center}
\end{figure}

For this task, we select a random class and remove it from the train dataset. Thereafter, we try to predict that class during testing from the set of all classes, including both novel and seen classes. It means that the classifier has to choose a class from among the 60 different classes present in the dataset. This is the hardest task and also closely resembles the real world situation where we do not know in advance whether the instance belongs to a seen or unseen class. The results are presented in Figure \ref{fig:mean_acc} and are statistically significant using a t-test at p-value $<$ 0.001. NSM-PB outperforms the baselines in this task as well (see Figure \ref{fig:mean_acc}).  We make the classifier choose from the complete set of 60 classes instead of restricting the possible set of classes. Note that we outperform  LM-PB, LM, NSM and ConSE by a significant margin in terms of classification accuracy (See Figure \ref{fig:mean_acc}). These results are as expected given that we minimize generalization error for nearest neighbor classifier. This also clearly shows in the results as both LM and ConSE  fall short in comparison to NSM for classification accuracy. These results clearly give us an insight regarding the usefulness of property embeddings, as LM-PB performs much better than LM and NSM-PB performs much better than NSM. In addition to property embeddings, the neighborhood sensitive mapping further improves the accuracy of classification.

\subsection{Text Classification Datasets}
After the initial experiments on fMRI dataset, we tested the best performing NSM-PB against LM and ConSE on two larger text classification datasets. \textit{First}, \textbf{wiki10}+ dataset \cite{zubiaga2012enhancing} contains text of \textit{wikipedia} articles and the tags assigned by users on \textit{delicious.com} for \emph{url} of those articles. We used the most popular tag for an article as the label of that article. There were 20762 instances in the dataset with 5303 distinct labels. We cleaned the data of all html tags and computed tf-idf representations of the data. Afterwards, we used truncated-SVD to reduce noise in the data.  \textit{Second}, \textbf{delicious} dataset \cite{tsoumakas2008effective} contains features of web pages from all over the internet with tags generated by users on those web pages as the labels. The dataset contains 500 features for each instance and 983 unique labels. It has more than 16000 instances and the features are binary, with $1$ indicating the presence of the feature and $0$ indicating the absence of the feature.

\subsection{Experiments}

\begin{table}[htb]
\begin{center}
\begin{tabular}{ c | c | c | c | c }
  K & NSM-PB & LM & ConSE & SSE\\
  \hline 
  5 & \textbf{0.0685} & 0.0261 & 0.0283 & 0.0269 \\
  \hline
  10 & \textbf{0.1344} & 0.0392 & 0.0472 & 0.0481\\
  \hline
  50 & \textbf{0.3590} & 0.0521 & 0.0825 & 0.1541\\
\end{tabular}
\end{center}
\caption{The value of accuracy for different methods on wiki10+ dataset. A given prediction is considered accurate if top-K labels in the prediction list contain the correct label. Results are statistically significant using t-test at p-value $<$ 0.001.}
\label{ref:table1}
\end{table}

\begin{table}[htb]
\begin{center}
\begin{tabular}{ c | c | c | c | c }
 K & NSM-PB & LM & ConSE & SSE \\
  \hline 
  5 & \textbf{0.0509} & 0.0320 & 0.0433 & 0.0274 \\
  \hline
  10 & \textbf{0.0924} & 0.0492 & 0.0822 & 0.0430\\
  \hline
  20 & \textbf{0.1486} & 0.0721 & 0.1350 & 0.0597\\
\end{tabular}
\end{center}
\caption{The value of accuracy for different methods on delicious dataset. A given prediction is considered accurate if top-K labels in the prediction list contain the correct label. Results are statistically significant using t-test at p-value $<$ 0.001.}
\label{ref:table2}
\end{table}

In this section we compare our approach against an additional baseline called SSE \cite{zhang2015zero}, which was a more recent method for zero-shot classification. 

We decided to test the accuracy of the methods using the top-K predicted labels, which means that the prediction was considered accurate if the top-K predicted labels contained the correct label.  We can see in Table \ref{ref:table1} that NSM-PB outperforms the other approaches by a considerable margin for varying levels of $K$. We can see that NSM-PB performed well even for $K=5$, which is a very small value considering that the classifier has more than 5000 labels to choose from. We can see very similar results in Table \ref{ref:table2} that NSM-PB again outperforms all other approaches for varying values of $K$.

\subsection{Runtime}
We observed that the running time for LM, ConSE and NSM were comparable, but the running time for SSE was much larger. The larger running time in the case of SSE can be attributed to sequentially learning label embeddings for all the labels in terms of proportions of seen labels. Therefore, SSE can take significant execution time when the seen label set contains a large number of labels.

\subsection{Reproducibility}
We share the codes used for the given experiments at \url{http://bit.ly/1RCNlwR}. The values of $\alpha$ and $\lambda$ (Equation \ref{eq:j11}) were tuned over a validation set, and were found to be 0.1 and 0.5 respectively. In case of ConSE, we use a multi-class logistic regression classifier for predicting class probabilities. The values of parameter T (i.e. number of top-T nearest embeddings for a given instance) in ConSE that gave best result was 5. The dimensionality of the learned property embeddings in the experiments was 10. The code for SSE was obtained from the matlab demo shared by the authors on their web-page. 

\section{Conclusion}

We proposed a model that uses independently learned semantic embeddings to improve a zero-shot text classifier in a multi-class classification setting. We first describe the issues associated  with zero-shot classification systems that use semantic embeddings. Then we highlight the shortcomings of classifiers that are not sensitive to the neighborhood of a label in the semantic space.  After that we show that our proposed data-dependent property embeddings combined with neighbourhood sensitive mapping leads to improved results for zero-shot classification in text. These novel embeddings reduce noise in the pre-trained semantic  embeddings, the proposed classification model then uses these newly learned embeddings for neighborhood-aware nearest neighbour classification. 
\bibliographystyle{splncs}
\bibliography{biblio}
\end{document}